\documentclass{article}

    \PassOptionsToPackage{numbers, compress}{natbib}
 \usepackage[preprint]{neurips_2026}

\usepackage[utf8]{inputenc} 
\usepackage[T1]{fontenc}    
\usepackage{hyperref}       
\usepackage{url}            
\usepackage{booktabs}       
\usepackage{amsfonts}       
\usepackage{nicefrac}       
\usepackage{microtype}      
\usepackage{xcolor}         

\usepackage{graphicx}
\usepackage{amsmath}
\usepackage{orcidlink}
\usepackage{makecell}
\usepackage{multirow}
\usepackage{cleveref}
\usepackage{algorithm}
\usepackage{algpseudocode}
\title{ProxyUp: Training-Free Proxy-Conditioned Video Generation for Controllable Dynamics}

\author{
	Zanwei Zhou$^{1,*}$\quad Jiazhong Cen$^{1,*}$\quad Jiemin Fang$^{2,\dagger}$\quad Yumeng He$^{1}$\quad Chen Yang$^{2}$\\ \textbf{Sikuang Li}$^{1}$\quad \textbf{Fanpeng Meng}$^{2}$\quad \textbf{Zhikuan Bao$^{2}$\quad Wei Shen$^{1,\dagger}$\quad Qi Tian$^{2,\dagger}$}\\
	$^{1}$Shanghai Jiao Tong University\quad $^{2}$Huawei Inc.\\
	\texttt{\small\{sjtu19zzw, jiazhongcen, ymhe, uranusits, wei.shen\}@sjtu.edu.cn}\\
	\texttt{\small \{jaminfong, chenyang.res, fpmeng0610, williambaozk\}@gmail.com} \;\;
	\texttt{\small tian.qi1@huawei.com} \\
	\url{https://zanue.github.io/proxyup}
}

\begin{document}

\maketitle

\begingroup
\renewcommand{\thefootnote}{} 
\footnotetext{$^*$ Equal contribution. Work done during internship at Huawei.}
\footnotetext{$^\dagger$ Corresponding author.}
\endgroup

\begin{abstract}
Precise control over complex dynamics remains challenging for modern video generative models, as text prompts alone often cannot specify physically plausible, fine-grained motion and interactions. We introduce \textit{proxy-conditioned video generation}, where a coarse proxy video from physics-based simulation or real-world recording serves as a dynamics carrier to control foreground object motion. Given a proxy video and a text prompt, the goal is to synthesize a new video that preserves the proxy dynamics while generating novel content and plausible interactions aligned with the prompt. Since paired proxy-target videos are difficult to obtain, we propose \textbf{ProxyUp}, a training-free framework built on pretrained video generative models. ProxyUp first inverts the proxy video into an intermediate latent representation and applies \textbf{region-wise latent noising}, preserving motion-critical proxy latents while injecting noise into regions intended for text-driven regeneration. To mitigate the distribution mismatch and weak foreground-background coupling introduced by this heuristic latent composition, we further propose \textbf{Stochastic Flow Relaxation (SFR)}, which progressively relaxes the composed latent toward the model's learned distribution before ODE sampling. Experiments on both simulation and real-world proxies show that ProxyUp outperforms strong video editing and motion transfer baselines in dynamic fidelity and text alignment.
\end{abstract}

\section{Introduction}
\label{sec:intro}

\begin{figure}[tbp]
    \centering
    \includegraphics[width=\linewidth]{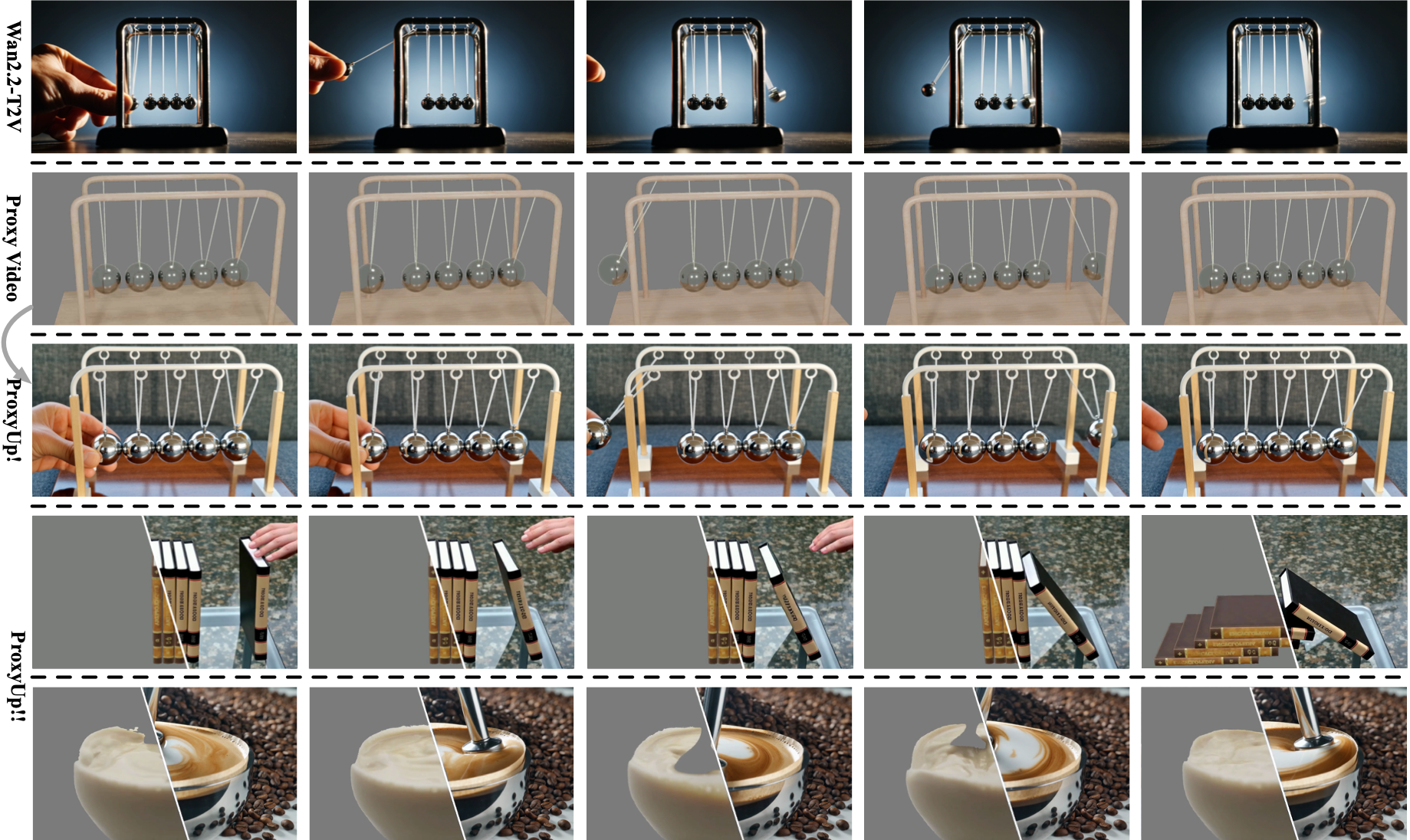}
    \vspace{-8pt}
    \caption{Given a proxy video (middle) as a dynamics carrier, \textbf{ProxyUp} generates videos that better follow the desired motion and interaction patterns compared with state-of-the-art video generation models such as Wan2.2 (top), while allowing flexible content regeneration.}
    \label{fig:teaser}
    \vspace{-28pt}
\end{figure}

Video generative models~\cite{hongcogvideo,yang2024cogvideox,wan2025wan,kong2024hunyuanvideo,blattmann2023stable,brooks2024video} have made rapid progress in visual quality and motion realism. 
However, precise control over complex temporal dynamics remains challenging. 
While text prompts provide an effective interface for specifying semantic content, they are often insufficient to describe fine-grained motion, object interactions, and complex dynamic processes. 
As a result, generated videos frequently exhibit implausible motion patterns, inconsistent temporal transitions, or weak coupling between interacting objects, as illustrated in Fig.~\ref{fig:teaser}. 
Such limitations hinder the use of current models in applications requiring controllable and reliable dynamics.

To address this issue, we explore an alternative setting: \emph{proxy-conditioned video generation}, where a proxy video serves as an explicit dynamics carrier. 
The proxy video can be a coarse, readily obtainable signal from low-fidelity physics simulation or real-world recording, while still providing structured temporal cues that are difficult to express with text alone.
Unlike conventional video editing, where the input video is treated as the target to be modified, the proxy video here is not an appearance reference but a source of dynamics, specifying the motion of objects over time while leaving content like interaction and appearance to be specified by the text prompt.

A natural solution is to train a model that takes both proxy videos and text prompts as input. 
However, this requires paired proxy-target videos that share similar dynamics but differ in appearance, object category, or scene context, which are difficult to obtain at scale. 
Moreover, existing alternatives do not fully address this setting. 
Motion transfer methods are mainly designed to transfer global motion from the source to the target, and therefore struggle to generate complex interactive background dynamics.
Video editing methods, on the other hand, are often anchored to the source video's appearance and layout, limiting their ability to generate substantially new content.
Inpainting methods complete masked regions from surrounding visible content, but object-environment interactions often occlude parts of the object, leaving them unable to generate the hidden content needed for coherent dynamics.
Neither directly supports the goal of using proxy videos purely as condition signals for controllable generation.

In this work, we propose \textbf{ProxyUp}, a training-free framework for proxy-conditioned controllable video generation. 
Given a proxy video and a text prompt, ProxyUp synthesizes a new video that follows the proxy's essential dynamics while regenerating content according to the prompt. 
Specifically, ProxyUp first performs ODE inversion on the proxy video to obtain an intermediate latent representation, and then applies \textbf{region-wise latent noising}: it preserves inverted latents in motion-critical regions (\emph{e.g.}, foreground objects) to retain key dynamics, while injecting Gaussian noise into regions designated for regeneration (\emph{e.g.}, background) to enable prompt-driven synthesis. 
However, such hybrid latent composition introduces distribution mismatch and weak cross-region coupling, leading to inconsistent interactions and degraded visual quality. 
To address this issue, we introduce \textbf{Stochastic Flow Relaxation (SFR)}, a multi-round procedure that progressively relaxes the composed latent toward the model's learned manifold before standard ODE sampling. 
Together, region-wise latent noising and SFR enable effective dynamics preservation while maintaining global coherence and visual fidelity.

To evaluate this new setting, we construct a task-specific evaluation set with both physics-simulation videos and real-world recordings as proxy videos. 
Existing video generation, editing, and motion-transfer benchmarks do not directly cover this setting, as they typically do not pair a proxy video with a target prompt. Such pairs are essential for evaluating whether a method can preserve proxy dynamics while regenerating prompt-specified content. 
Extensive experiments on this evaluation set demonstrate that ProxyUp consistently outperforms strong video editing and motion transfer baselines in dynamic fidelity and text alignment.
We hope this work provides a practical approach to controllable video generation with complex dynamics and inspires future research on leveraging proxy signals for video synthesis.

In summary, our contributions are as follows:
\begin{itemize}
    \item We introduce \textit{proxy-conditioned video generation}, a new setting where proxy videos act as dynamics carriers while text prompts specify the interactive scene context.
    \item We propose \textbf{ProxyUp}, a training-free framework that combines region-wise latent noising and stochastic flow relaxation to preserve motion-critical proxy dynamics while enabling prompt-driven content regeneration and coherent foreground-background interactions.
    \item We construct a task-specific evaluation set with both \textbf{simulation-based} and \textbf{real-world} proxy videos, and demonstrate that ProxyUp improves dynamic fidelity and text alignment over representative video editing and motion-transfer baselines.
\end{itemize}

\section{Related Work}

\subsection{Controllable Video Generation} Prior work on controllable video generation introduces various conditioning signals beyond text and images, including human motion cues (2D/3D skeletons, SMPL parameters)~\cite{ma2024follow, zhong2024posecrafter,zhou2026few}, explicit motion representations (optical flow, point trajectories)~\cite{wang2023videocomposer, nam2025optical}, structural priors (edges, depth, normals)~\cite{chen2023control, hu2023videocontrolnet, wang2023videocomposer}, semantic layouts (segmentation maps, mask tubes, bounding boxes)~\cite{li2025trackdiffusion, kim2025target, xi2025omnivdiff}, and camera controls (extrinsics, intrinsics, view trajectories)~\cite{he2024cameractrl, bai2025recammaster, bai2024syncammaster, cao2025uni3c}. More recent efforts further explore physics-related conditions such as contact cues, state constraints, and simulation rollouts to improve interaction plausibility.

\subsection{Motion Transfer}
Motion transfer aims to synthesize a target video whose motion follows a driving source while its appearance or identity is changed. Early systems often rely on explicit intermediate representations, such as human pose or learned keypoints, to retarget motion from a source performer or object to a target subject~\cite{chan2019everybody, siarohin2019first}. Recent diffusion-based methods extend this idea to text-conditioned video synthesis: MotionDirector~\cite{zhao2024motiondirector} learns reusable motion concepts with decoupled spatial and temporal LoRAs, MotionFlow~\cite{meral2026motionflow} transfers motion through attention-derived masks in pretrained video diffusion models, and DiTFlow~\cite{ditflow} extracts attention motion flow from diffusion transformers for training-free motion transfer. These methods provide strong mechanisms for copying visual motion patterns from reference videos, but they typically assume the reference video is itself a source of transferable visual motion or subject behavior. In contrast, our goal is not merely to retarget an observed appearance-motion pair, but to preserve motion-critical dynamics while regenerating content and interactions under a new prompt.

\subsection{Video Editing and Inpainting}
Diffusion-based~\cite{ddpm, song2020score0based, flowmatching} video editing can be roughly grouped by the underlying generative model. Early methods mainly adapt text-to-image models~\cite{rombach2021high0resolution, imagen, sd3.5, podell2023sdxl0} to video through techniques such as inversion, feature injection, and cross-frame regularization~\cite{ceylan2023pix2video,vid2vid-zero,khachatryan2023text2video,fatezero,tokenflow,flatten,rave,wu2023tune,liu2024video,molad2023dreamix}. More recent approaches leverage native video generation models~\cite{make-a-video, hongcogvideo, vdm, blattmann2023stable, svd, animatediff, kong2024hunyuanvideo, wan2025wan, ModelScope, MagicVideo-v2, VideoCrafter1} and their spatiotemporal priors for improved temporal consistency~\cite{videoswap,videodirector,ouyang2024i2vedit,chen2025contextflow,kim2025tv}. Video inpainting methods instead focus on completing masked spatiotemporal holes using visible context, from flow- or transformer-based propagation~\cite{li2022e2fgvi,liu2021fuseformer,zhou2023propainter} to unified masked video creation and editing~\cite{vace}. While editing and inpainting provide useful tools for modifying or completing input videos, their objectives preserve the input structure or visible regions rather than use an external video purely as a dynamics prior for prompt-driven generation. In this work, we draw inspiration from the representative diffusion-based editing paradigm SDEdit~\cite{sdedit} and build a method for our proxy guided video generation setting.

\section{Preliminary Analysis: Editing, Inpainting, and Motion Transfer with Proxy Videos}

To transfer dynamics from a proxy video to a new visual context, a natural starting point is to use existing video editing, inpainting, or motion transfer methods~\cite{sdedit, flowedit, li2025flowdirector, vace}. These methods are designed to preserve certain input cues while changing the generated appearance. However, they expose a key trade-off in our setting: stronger preservation keeps the result too close to the proxy video's original, often simplified appearance, while stronger modification makes it difficult to retain the underlying dynamics.

\begin{figure}[tbp]
    \centering
    \includegraphics[width=\linewidth]{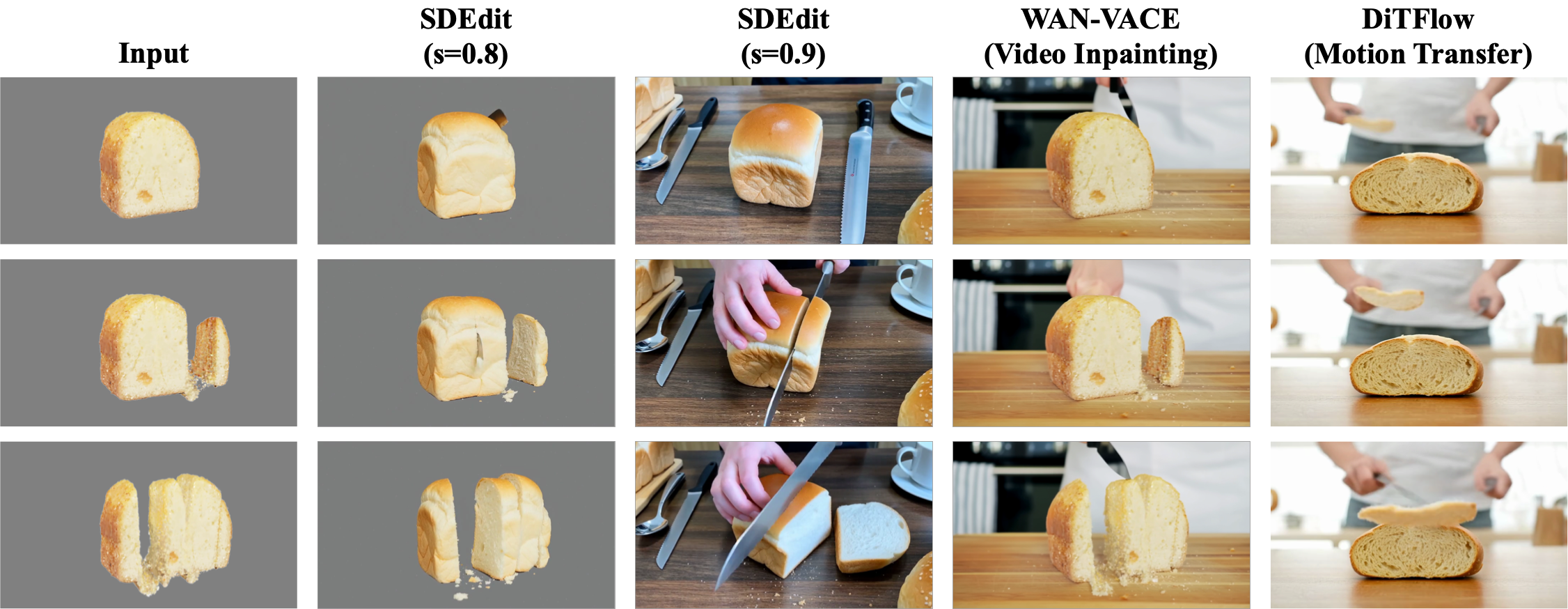}
    \vspace{-8pt}
    \caption{Preliminary analysis of proxy-video-guided generation using existing pipelines.
    We evaluate representative video editing, inpainting and motion transfer baselines as straightforward approaches for transferring dynamics from a proxy video. The text prompt describes a person using a knife to cut bread in a kitchen.}
    \label{fig:preliminary}
    \vspace{-12pt}
\end{figure}

We examine this limitation using representative pipelines in Fig.~\ref{fig:preliminary}. For video editing, SDEdit preserves some motion cues from the proxy video, but struggles to generate text-aligned content with coherent dynamics. When the editing strength\footnote{We use strength to denote the ratio of the remaining sampling steps at the time $t$ to the total number of inference steps.} $s$ is set to 0.8 (corresponding to $t=0.922$ in our sampling scheduler), the result still inherits the gray proxy background. Increasing $s$ to 0.9 ($t=0.964$) allows larger appearance changes, but largely removes the original dynamics. This behavior reflects the objective of editing methods: they are mainly designed for local content modification, so the output remains strongly anchored to the input video.

Inpainting-based methods provide more flexibility in masked regions, but their task formulation is also misaligned with our goal. They complete missing content while preserving visible regions, rather than using the proxy video as a dynamics prior for synthesizing a new, physically coherent scene. For example, in Fig.~\ref{fig:preliminary} VACE keeps the foreground closely aligned with the proxy video, but fails to modify it into a coherent interaction between the knife and the bread.

Motion transfer methods are closer to our objective because they explicitly transfer motion from a reference video. However, representative methods such as DiTFlow~\cite{ditflow} rely on frame- and location-wise correspondences between the source and target videos. This one-to-one formulation can reuse existing motion patterns, but cannot easily synthesize action components that are absent from the proxy video yet necessary in the target scene. As shown in Fig.~\ref{fig:preliminary}, DiTFlow struggles to produce a coherent bread-cutting interaction because the required human-object interaction is not directly present in the proxy video.

These observations suggest that existing editing, inpainting, and motion transfer frameworks are not sufficient for our setting. This motivates us to introduce \textbf{ProxyUp}, a dedicated framework that explicitly preserves dynamics from the proxy video while enabling prompt-driven synthesis of new content.

\section{Method}

\begin{figure}[tbp]
    \centering
    \includegraphics[width=\linewidth]{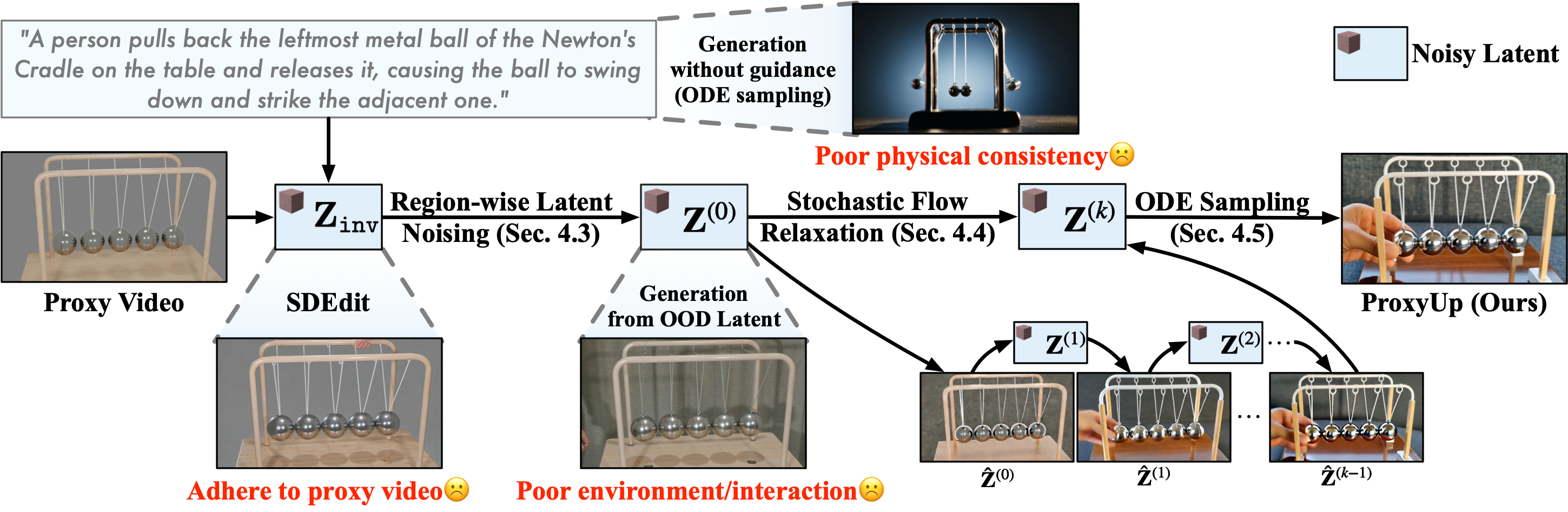}
    \vspace{-8pt}
    \caption{Overview of ProxyUp. Given a proxy video as guidance, we first apply Region-wise Latent Noising to preserve the foreground dynamics while ensuring the generative flexibility of background environment. We then perform Stochastic Flow Relaxation (SFR) to refine the hand-made out-of-distribution latent for better visual quality and foreground-background coherence. Finally, an ODE solver is applied to generate the final video.}
    \label{fig:main}
    \vspace{-8pt}
\end{figure}

In this section, we present \textbf{ProxyUp}, a training-free framework that leverages proxy videos to guide high-fidelity video generation. An overview of our approach is illustrated in Fig.~\ref{fig:main}. ProxyUp operates by first constructing a motion-preserving latent, refining it via a stochastic relaxation process to ensure spatial-temporal coherence, and finally performing deterministic sampling to produce the guided video.

\subsection{Preliminaries: Rectified Flow}
\label{sec:prelim}

Rectified Flow~\cite{rectifiedflow} learns a probability flow ordinary differential equation (ODE) that transforms a noise distribution into a data distribution by modeling a time-dependent velocity field. Let $\mathbf{Z}_0 \sim p_{\text{data}}$ denote a clean sample and $\mathbf{Z}_1 \sim \mathcal{N}(\mathbf{0}, \mathbf{I})$ denote Gaussian noise. Rectified Flow induces a linear interpolation path:
\begin{equation}
    \mathbf{Z}_t = (1-t)\mathbf{Z}_0 + t\mathbf{Z}_1, \quad t \in [0,1].
\end{equation}
A neural network $\mathbf{v}_\theta(\mathbf{Z}_t, t, c)$ is optimized to predict the constant velocity of this trajectory, conditioned on a signal $c$. The training objective simplifies to a regression of the target displacement:
\begin{equation}
    \mathcal{L}_{\mathrm{RF}} = \mathbb{E}_{\mathbf{Z}_0, \mathbf{Z}_1, t}\left[\left\| \mathbf{v}_\theta(\mathbf{Z}_t, t, c) - (\mathbf{Z}_1 - \mathbf{Z}_0) \right\|_2^2\right].
\end{equation}

At inference, generation is performed by solving the learned ODE from $t=1$ to $t=0$. Using a first-order Euler solver, the latent is updated as:
\begin{equation}
    \mathbf{Z}_{t-\Delta t} = \mathbf{Z}_{t} - \Delta t \cdot \mathbf{v}_\theta(\mathbf{Z}_{t}, t, c),
\end{equation}
where $\Delta t > 0$ is the step size. 

\subsection{Problem Formulation and Pipeline Overview}
\label{sec:overview}

Given a proxy video $\mathcal{V}_{p}$, a foreground mask $\mathbf{M}$ that specifies the region whose dynamics should be retained, and a text prompt $c$, our goal is to synthesize a high-fidelity video $\mathcal{V}_{g}$ that follows the dynamics of the masked foreground in $\mathcal{V}_{p}$ while generating a novel environment which interacts with the foreground and is consistent with the text prompt. The central challenge is to balance two competing requirements: the model must faithfully follow the dynamics provided by the masked foreground prior, while retaining sufficient generative flexibility to avoid inheriting the original appearance of the proxy video.

To address this challenge, we propose \textbf{ProxyUp}, a three-stage framework:
\begin{enumerate}
    \item \textbf{Region-wise Latent Noising (Sec.~\ref{sec:noising}):} To preserve dynamic guidance in the masked foreground region while granting the model sufficient freedom to synthesize a new environment, we introduce a region-wise latent construction strategy that retains motion-guiding latents within $\mathbf{M}$ and leave sufficient space to generate a corresponding background environment outside $\mathbf{M}$.
    \item \textbf{Stochastic Flow Relaxation (Sec.~\ref{sec:sfr}):} Although this region-wise construction balances dynamic guidance and generative flexibility, it inevitably produces an \textit{out-of-distribution (OOD)} latent with weak coupling between the masked foreground and the generated background. We therefore propose SFR to iteratively relax this hybrid latent toward the model's learned in-distribution manifold.
    \item \textbf{Guided ODE Sampling (Sec.~\ref{sec:sampling}):} Finally, we perform deterministic ODE sampling from the relaxed latent to generate the output video.
\end{enumerate}

\subsection{Region-wise Latent Noising for Dynamic Guidance}
\label{sec:noising}

To flexibly generate an environment corresponding to the text prompt while maintaining the dynamics of the foreground, we must decouple the motion of the reference object from its original environment. Given the latent representation $\mathbf{Z}_0$ of the proxy video and a binary mask $\mathbf{M}$ indicating the guidance region, we first isolate the foreground as $\mathbf{M} \odot \mathbf{Z}_0$. We then perform an ODE-based inversion on this masked latent to an intermediate noise level $t_{\text{init}}$, yielding a structured noisy representation $\mathbf{Z}_{inv}$. This inversion preserves the motion layout and temporal dynamics of the original foreground while preventing leakage from the proxy background.

We then construct an initial mixed latent $\mathbf{Z}^{(0)}$ at timestep $t_{\text{init}}$. For the background region, instead of using the inverted proxy background, we inject noise $\epsilon_{bg}$ that matches the marginal distribution at $t_{\text{init}}$:
\begin{equation}
    \epsilon_{bg} \sim \mathcal{N}\big(\mathbf{0}, ((1-t_{\text{init}})^2 + t_{\text{init}}^2)\mathbf{I}\big).
\end{equation}
The mixed latent is then composed as:
\begin{equation}
    \mathbf{Z}^{(0)} = \mathbf{M} \odot \mathbf{Z}_{inv} + (1 - \mathbf{M}) \odot \epsilon_{bg}.
\end{equation}
This design ensures the model adheres to the guiding motion priors while retaining maximum generative flexibility to synthesize a new environment conditioned on the text prompt $c$.

\subsection{Stochastic Flow Relaxation (SFR)}
\label{sec:sfr}

While Region-wise Latent Noising preserves desired dynamics, the resulting mixed latent $\mathbf{Z}^{(0)}$ is a hand-made construction that lacks the inter-region spatiotemporal coupling learned by the flow matching model. Direct sampling from this latent will produce artifacts such as unnatural boundaries or implausible interactions. To address this, we propose \textbf{Stochastic Flow Relaxation (SFR)}, which improves the generation quality by iteratively refining the latent to match the marginal distribution modeled by the flow matching model at the fixed noise level $t_{\text{init}}$.

Specifically, for $K$ iterations, we perform a two-stage stochastic update:
\begin{enumerate}
    \item \textbf{Flow-induced denoising:} We estimate the clean latent $\hat{\mathbf{Z}}_0^{(k)}$ using a one-step Euler update toward the data direction:
    \begin{equation}
        \hat{\mathbf{Z}}_0^{(k)} = \mathbf{Z}^{(k-1)} - t_{\text{init}} \cdot \mathbf{v}_\theta(\mathbf{Z}^{(k-1)}, t_{\text{init}}, c).
    \end{equation}
    \item \textbf{Re-noising:} We inject noise $\epsilon^{(k)} \sim \mathcal{N}(0, \mathbf{I})$ to map the latent back to the noise level $t_{\text{init}}$:
    \begin{equation}
        \mathbf{Z}^{(k)} = (1 - t_{\text{init}}) \cdot \hat{\mathbf{Z}}_0^{(k)} + t_{\text{init}} \cdot \epsilon^{(k)}.
    \end{equation}
\end{enumerate}

This cycle enables the model to significantly reduce OOD artifacts, foster more natural interactions between the foreground and the generated background, and ensure the environment aligns faithfully with the text prompt. In fact, as the cycle iteration $K$ increase, we can conclude the following:

\vspace{2mm}
\noindent\fbox{
    \parbox{0.96\linewidth}{
        \textbf{Proposition 1 (Stochastic Flow Relaxation).} \textit{Given an OOD latent $\mathbf{Z}^{(0)}$, the SFR procedure defines a Markov refinement process that progressively aligns the latent distribution with the flow matching model's learned in-distribution prior, driving it toward the learned video manifold.}
    }
}
\vspace{2mm}

Please refer to the appendix~\ref{sec:proof_proposition_1} for the detailed proof. Based on this proposition, when $K$ is sufficiently large, the distribution of the noisy latent $\mathbf{Z}^{(k)}$ can approach the marginal distribution modeled by flow matching at time $t_{\text{init}}$, thereby ensuring the diversity and visual quality of the generated videos.

\subsection{ODE Solver for Deterministic Generation}
\label{sec:sampling}

After the latent has been sufficiently refined via SFR, we proceed with standard deterministic generation. Starting from the relaxed latent $\mathbf{Z}^{(K)}$ at $t_{\text{init}}$, we solve the learned ODE downward to $t=0$:
\begin{equation}
    \mathbf{Z}_{t_{i+1}} = \mathbf{Z}_{t_i} + \Delta t \cdot \mathbf{v}_\theta(\mathbf{Z}_{t_i}, t_i, c),
\end{equation}
where $\Delta t$ is the discretization step. Because the starting latent $\mathbf{Z}^{(K)}$ is well-aligned with the model's prior, the sampling process reliably generates a foreground-background coherent result. The final latent $\mathbf{Z}_0$ is then decoded into the pixel-space video.

\section{Experiments}

\subsection{Experiment Settings}

We build ProxyUp on Wan2.2~\cite{wan2025wan} and evaluate it on a task-specific proxy-guided generation set with 76 proxy videos, split evenly between simulation-based and real-world proxies. The evaluation covers visual quality, proxy-dynamics fidelity, interaction plausibility, and material consistency using the IQ, MR, Mech., and Mat. dimensions. We summarize only the essential protocol here; full implementation details, dataset construction, and metric definitions are provided in Appendix~\ref{sec:appendix_exp_details}.



\begin{table}[t]
\centering
\caption{Quantitative comparison on the real-video and physics-simulation subsets.}
\label{tab:quantitative}
\setlength{\tabcolsep}{2.5pt}
\begin{tabular}{llcccccc}
\toprule
Subset & Metric & Wan2.2 & VACE & DiTFlow & SDEdit & FlowDirector & ProxyUp \\
\midrule
\multirow{4}{*}{Real} & IQ & 66.39 & 63.08 & 64.11 & 61.49 & 66.40 & \textbf{68.67} \\
 & MR & 0.70 & 0.65 & 0.57 & 0.48 & 0.65 & \textbf{0.73} \\
 & Mech. & 0.91 & 0.87 & 0.72 & 0.88 & 0.87 & \textbf{1.00} \\
 & Mat. & 0.80 & 0.88 & 0.60 & 0.86 & 0.73 & \textbf{0.89} \\
\midrule
\multirow{4}{*}{Sim.} & IQ & 64.42 & 62.89 & 57.18 & 67.99 & 64.12 & \textbf{68.88} \\
 & MR & 0.31 & 0.41 & 0.27 & 0.28 & 0.45 & \textbf{0.52} \\
 & Mech. & 0.86 & 0.92 & 0.75 & 0.96 & 0.90 & \textbf{1.00} \\
 & Mat. & \textbf{1.00} & \textbf{1.00} & 0.50 & \textbf{1.00} & 0.89 & \textbf{1.00} \\
\bottomrule
\end{tabular}
\vspace{-10pt}
\end{table}

\subsection{Quantitative Results}

In Table~\ref{tab:quantitative}, we evaluate ProxyUp on the real-video and physics-simulation test sets, respectively. Wan2.2~\cite{wan2025wan} is a text-only baseline with state-of-the-art performance. VACE~\cite{vace} is a video inpainting method, DiTFlow~\cite{ditflow} is a training-free motion transfer method, while FlowDirector~\cite{li2025flowdirector} is a training-free video editing method; both are conditioned on a video and a text prompt. SDEdit~\cite{sdedit} is a diffusion-based image editing method that can be naturally extended to video diffusion models. We also compare with VACE~\cite{vace}, which provides a video inpainting baseline. The implementation details of all baselines are provided in Appendix~\ref{sec:appendix_baseline_implementations}.

Except for Wan2.2, all baselines take the proxy video and text prompt as input. Across both benchmarks, ProxyUp achieves the best dynamics consistency on dynamics-related metrics without compromising imaging quality, with clear gains on Motion Rationality (MR) and Mechanics (Mech.).

\begin{figure}[tbp]
    \centering
    \includegraphics[width=\linewidth]{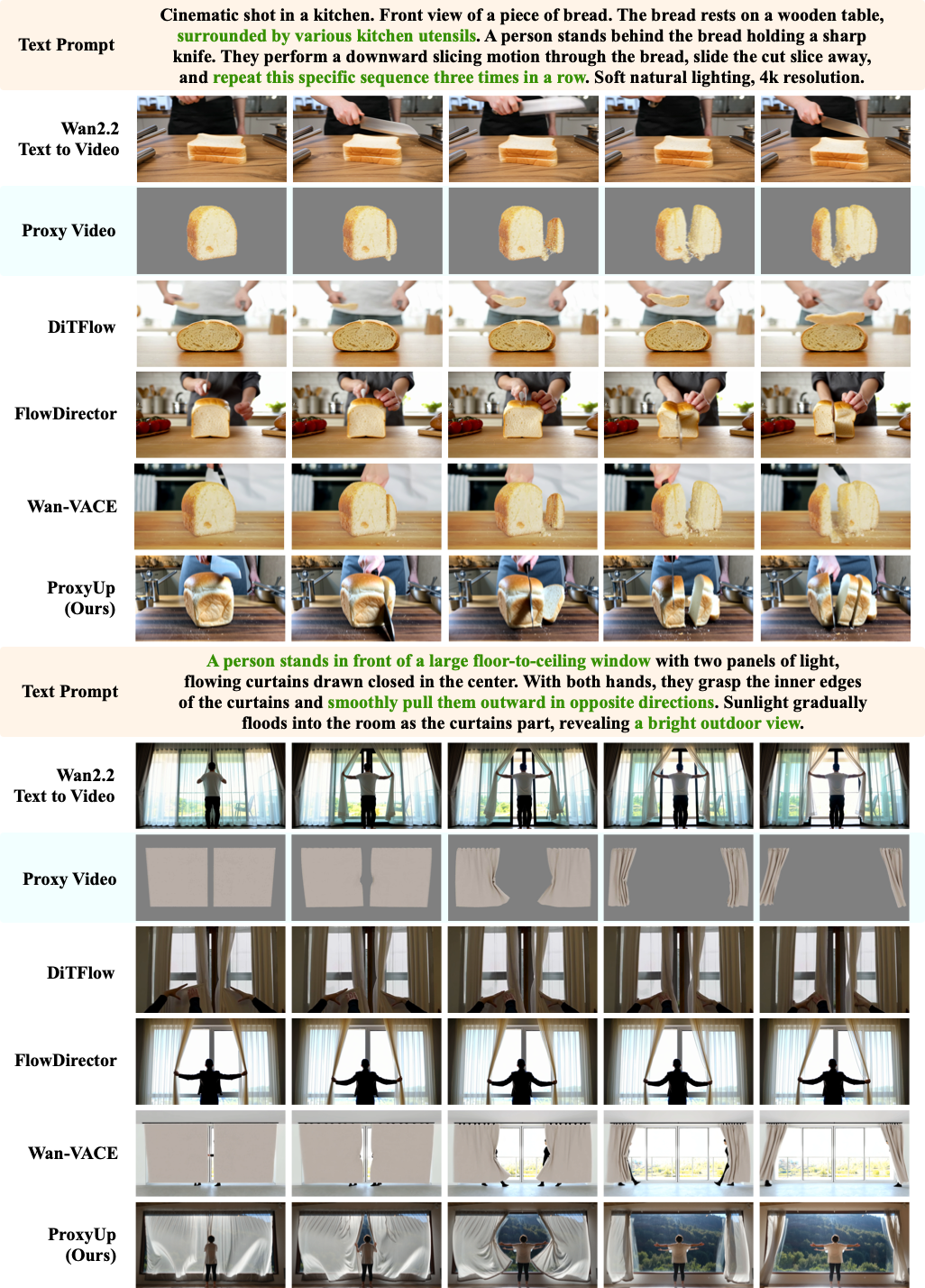}
    \caption{Qualitative comparison results. Methods presented below the proxy video use the same proxy video as conditioning input. Existing methods struggle to accurately follow the text prompt and synthesize the intended interaction and physical phenomenon from the proxy video.}
    \label{fig:qualitative_comp}
\end{figure}

\subsection{Qualitative Results}

As shown in Fig.~\ref{fig:qualitative_comp}, we qualitatively compare ProxyUp with four representative baselines. The proxy videos adopted by ProxyUp can be obtained from a wide range of sources, including physics simulation and graphics rendering. In this figure, the first proxy video (cutting bread) comes from an MPM simulation, whereas the second proxy video (pulling curtains) is rendered in Blender.

ProxyUp captures proxy dynamics and translates them into more accurate, prompt-aligned interactions. In bread cutting, other methods fail to follow the specified three repeated cuts, whereas ProxyUp better reproduces the motion with more natural knife--bread interaction and deformation. Similarly, in curtain pulling, DiTFlow, FlowDirector and VACE fail to generate the intended dynamics even with proxy guidance.

\begin{figure}[tbp]
    \centering
    \includegraphics[width=\linewidth]{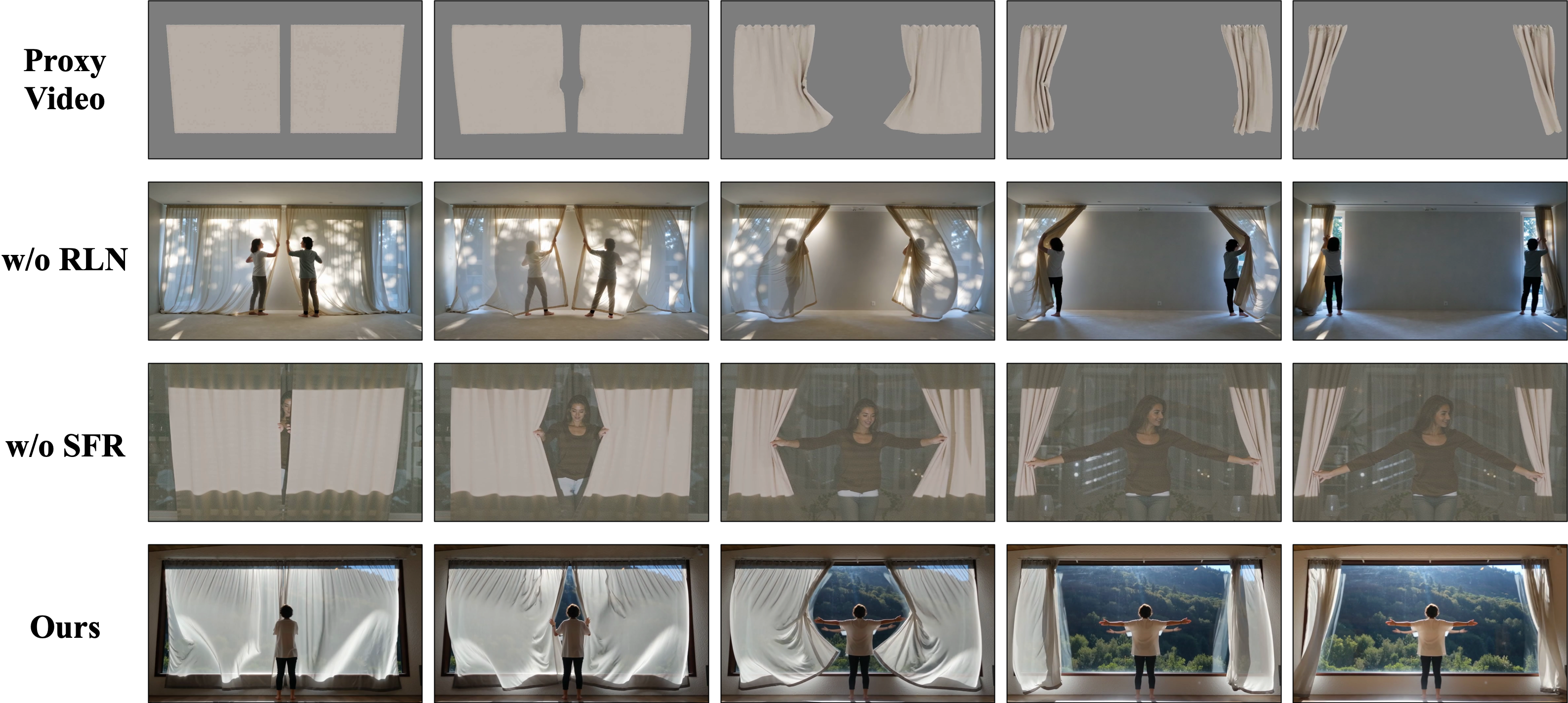}
    \vspace{-8pt}
    \caption{Ablation study on the key components of ProxyUp. Without region-wise latent noising (RLN), the output is overly tied to the proxy appearance and struggles to synthesize new content (\emph{e.g.}, outdoor scenes). Without stochastic flow relaxation (SFR), visual quality and content richness degrade. The prompt is the same as in Fig.~\ref{fig:qualitative_comp}.}
    \label{fig:ablation_wo}
    \vspace{-14pt}
\end{figure}

\subsection{Ablation Study And Hyperparameter Analysis}

As shown in Fig.~\ref{fig:ablation_wo}, we analyze the contribution of each component in ProxyUp. Without region-wise latent noising (RLN), even though SFR helps maintain reasonable visual quality, the generated video remains overly tied to the proxy video's original appearance, leading to failures such as producing a gray wall instead of the intended outdoor scene. Without SFR, the generated video quality degrades substantially, with noticeable noise and blurry, incoherent details.

We further analyze the choice of hyperparameters in our framework, with generated results under different values of $K$ and $t_{\text{init}}$ shown in Appendix Fig.~\ref{fig:rs}. We observe that when $K$ is relatively small (\emph{e.g.}, $K=5$), the generated videos remain noisy and visually simplistic, and often fail to produce the desired physical actions. As $K$ increases, the generated frames exhibit richer colors and finer details. To balance dynamics preservation, visual fidelity, and inference cost, we choose $K=15$ as the default setting. When $t_{\text{init}}$ is relatively small, we observe that it is hard to synthesize a high-quality background and fails to transform the foreground into a semantically plausible object. As the strength $s$ increases, the foreground progressively morphs into a bottle that aligns with the text prompt, while the background becomes increasingly clear. To prevent excessive degradation of the foreground and preserve its underlying motion dynamics, we empirically adopt the setting of $s=0.8$.

\section{Limitation}
ProxyUp depends on both the quality of the proxy video and the generative prior of the underlying video model. When the proxy contains uncommon dynamics or ambiguous force application points, the model may fail to infer the intended interaction, even if the proxy provides useful motion cues. In addition, ProxyUp is training-free and therefore cannot introduce physical or semantic knowledge that is absent from the pretrained generator. These limitations suggest that future work may benefit from stronger proxy construction, explicit interaction annotations, or video models trained with broader physical interaction data. Representative failure cases are discussed in Appendix~\ref{sec:appendix_failure_cases}.

\section{Conclusion}
In this paper, we presented \textbf{ProxyUp}, a training-free framework for proxy guided video generation. By using proxy videos as explicit dynamics priors, ProxyUp enables video generators to preserve meaningful motion cues while synthesizing novel content specified by text prompts. By combining region-wise latent noising with \textbf{Stochastic Flow Relaxation (SFR)}, ProxyUp preserves physically consistent dynamics from the proxy video without inheriting its visual details. Extensive experiments demonstrate the effectiveness of ProxyUp.

\bibliographystyle{plain}
\bibliography{main}


\appendix

\section{Proof of Proposition 1}
\label{sec:proof_proposition_1}

In this section, we provide the theoretical proof for Proposition 1. We aim to show that for an out-of-distribution (OOD) input $\mathbf{Z}^{(0)}$, the iterative Stochastic Flow Relaxation (SFR) cycle strictly minimizes the Kullback-Leibler divergence between the current state distribution $q_K$ and the learned in-distribution prior $\pi_{ID}$ at the fixed time $t$, i.e., $\lim_{K \to \infty} D_{KL}(q_K \parallel \pi_{ID}) = 0$.

\textbf{Step 1: Formulating the Markov Transition Kernel.}
Assume the Flow Matching model is well-trained, predicting the optimal velocity field $\mathbf{v}_\theta(\mathbf{Z}, t)$ that targets the in-distribution data manifold. Based on the theory of flow matching and Tweedie's formula, the one-step denoising prediction $\mathbf{\hat{Z}}_0^{(k)}$ is mathematically equivalent to computing the posterior expectation of the clean data given the noisy state $\mathbf{Z}^{(k-1)}$:
\begin{equation}
    \mathbf{\hat{Z}}_0^{(k)} = \mathbf{Z}^{(k-1)} - t \cdot \mathbf{v}_\theta(\mathbf{Z}^{(k-1)}, t) = \mathbb{E}_{ID}[\mathbf{Z}_0 \mid \mathbf{Z}^{(k-1)}]
\end{equation}
Following the re-noising step $\mathbf{Z}^{(k)} = (1 - t) \cdot \mathbf{\hat{Z}}_0^{(k)} + t \cdot \epsilon^{(k)}$, the overall SFR loop forms a first-order time-homogeneous Markov Chain. The transition kernel $K(\mathbf{Z}' \mid \mathbf{Z})$ from step $(k-1)$ to step $k$ is an additive Gaussian kernel:
\begin{equation}
    K(\mathbf{Z}' \mid \mathbf{Z}) = \mathcal{N}\left(\mathbf{Z}'; (1-t) \mathbb{E}_{ID}[\mathbf{Z}_0 \mid \mathbf{Z}], t^2 \mathbf{I}\right)
\end{equation}

\textbf{Step 2: Strict Convergence of the KL Divergence.}
At iteration $k$, the distribution $q_k(\mathbf{Z}')$ is updated via the transition kernel:
\begin{equation}
    q_k(\mathbf{Z}') = \int K(\mathbf{Z}' \mid \mathbf{Z}) q_{k-1}(\mathbf{Z}) d\mathbf{Z}
\end{equation}
According to the Data Processing Inequality (DPI) for Markov chains, applying a transition kernel monotonically decreases the KL divergence between any two distributions. Since $\pi_{ID}$ is the invariant stationary distribution (i.e., $\pi_{ID} = \int K \pi_{ID}$), we have:
\begin{equation}
    D_{KL}(q_k \parallel \pi_{ID}) \le D_{KL}(q_{k-1} \parallel \pi_{ID})
\end{equation}
Since the sequence of KL divergence is non-negative and monotonically decreasing, it must converge to a limit. 

Crucially, because the re-noising step injects Gaussian noise with a strictly positive variance $t^2 \mathbf{I}$, the transition density satisfies $K(\mathbf{Z}' \mid \mathbf{Z}) > 0$ everywhere. This everywhere-positivity ensures that the Markov chain is ergodic. For such an ergodic chain with a strictly positive kernel, the equality in the DPI holds \emph{if and only if} $q_{k-1} = \pi_{ID}$. 

Therefore, the KL divergence strictly decreases at each step unless the distribution has already reached the unique stationary distribution $\pi_{ID}$. As $K \to \infty$, the distribution $q_K$ inevitably converges to this steady state, yielding:
\begin{equation}
    \lim_{K \to \infty} D_{KL}(q_K \parallel \pi_{ID}) = 0
\end{equation}
This completes the proof. It guarantees that the SFR loop irreversibly dissipates the initial OOD structural artifacts, pulling the latent onto the learned video manifold $\pi_{ID}$ before generation.

\section{Algorithm}

We present the complete implementation of our method in Algorithm~\ref{alg:ProxyUp}.

\begin{algorithm}[htbp]
\caption{ProxyUp: Proxy-Guided Video Generation}
\label{alg:ProxyUp}
\begin{algorithmic}[1]
\Require Flow matching model $\mathbf{v}_\theta$, proxy latent $\mathbf{Z}_0$, mask $\mathbf{M}$, Condition $c$, noise level $t_{\text{init}}$, SFR iterations $K$ and step size $\Delta t$.

\Statex \textit{// Stage 1: Region-wise Latent Noising}
\State $\mathbf{Z}_{fg} \gets \mathbf{M} \odot \mathbf{Z}_0$ \Comment{Extract foreground}
\State $\mathbf{Z}_{inv} \gets \text{ODE\_Invert}(\mathbf{Z}_{fg}, 0 \to t_{\text{init}})$ \Comment{Invert to $t_{\text{init}}$}
\State $\sigma_{bg}^2 \gets (1-t_{\text{init}})^2 + t_{\text{init}}^2$
\State $\epsilon_{bg} \sim \mathcal{N}(\mathbf{0}, \sigma_{bg}^2\mathbf{I})$ \Comment{Sample background noise}
\State $\mathbf{Z}^{(0)} \gets \mathbf{M} \odot \mathbf{Z}_{inv} + (1 - \mathbf{M}) \odot \epsilon_{bg}$ \Comment{Construct mixed latent}

\Statex \textit{// Stage 2: Stochastic Flow Relaxation (SFR)}
\For{$k = 1, 2, \dots, K$}
    \State $\hat{\mathbf{Z}}_0^{(k)} \gets \mathbf{Z}^{(k-1)} - t_{\text{init}} \cdot \mathbf{v}_\theta(\mathbf{Z}^{(k-1)}, t_{\text{init}}, c)$ \Comment{Flow-induced denoising}
    \State $\epsilon^{(k)} \sim \mathcal{N}(\mathbf{0}, \mathbf{I})$ \Comment{Sample noise}
    \State $\mathbf{Z}^{(k)} \gets (1 - t_{\text{init}}) \cdot \hat{\mathbf{Z}}_0^{(k)} + t_{\text{init}} \cdot \epsilon^{(k)}$ \Comment{Re-noising}
\EndFor

\Statex \textit{// Stage 3: Guided ODE Sampling}
\State $\mathbf{Z}_{t_{\text{init}}} \gets \mathbf{Z}^{(K)}$ \Comment{Initialize from relaxed latent}
\For{$t = t_{\text{init}}, t_{\text{init}}-\Delta t, \dots, \Delta t$}
    \State $\mathbf{Z}_{t - \Delta t} \gets \mathbf{Z}_t - \Delta t \cdot \mathbf{v}_\theta(\mathbf{Z}_t, t, c)$ \Comment{Deterministic Euler update}
\EndFor

\State $\mathcal{V}_g \gets \text{Decode}(\mathbf{Z}_0)$ \Comment{Decode to pixel space}
\State \Return $\mathcal{V}_g$
\end{algorithmic}
\end{algorithm}

\section{Additional Experimental Details}
\label{sec:appendix_exp_details}

\subsection{Implementation Details}
We build ProxyUp primarily on top of Wan2.2~\cite{wan2025wan}. Unless otherwise specified, all videos are generated at a resolution of $832 \times 480$. In all experiments, we use a shifted sampling scheduler with a shift value of 3.0. We use strength $s$ to denote the ratio of the remaining sampling steps at a specific time step $t$ to the total number of inference steps. We set the total number of inference steps to 40, the intermediate noise level $t_{\texttt{init}}$ to 0.922 ($s=0.8$ in our sampling scheduler), and the number of stochastic flow relaxation steps $K$ to 15. Therefore, the numbers of inverse steps and sampling steps are both $40 \times 0.8 = 32$. For classifier-free guidance (CFG)~\cite{cfg}, we set it to 1 during inversion and 3 during stochastic flow relaxation, and follow Wan2.2 during ODE sampling by using guidance scales of 4 and 3 for the high-noise and low-noise stages, respectively. Although these additional inference-time operations increase computational cost, ProxyUp remains a training-free method; compared with collecting massive video corpora and retraining a video generation model, this moderate overhead at inference time is a practical and acceptable trade-off. All experiments are conducted on 8 NVIDIA A100 GPUs using Fully Sharded Data Parallel (FSDP)~\cite{fsdp} with Ulysses sequence parallelism~\cite{ulysses}, both configured with size 8.

\subsection{Evaluation Datasets}
Existing video generation, editing, and motion-transfer benchmarks are not directly aligned with our proxy-guided setting, since they typically provide only text prompts or source videos, but do not jointly provide proxy videos, foreground masks, and target prompts for dynamics-preserving regeneration. Therefore, following the common practice in controllable video generation and video editing, we construct a task-specific evaluation set for proxy-guided dynamic generation. The set contains 76 proxy videos, including 38 simulation-based videos and 38 real-world videos. The simulation subset is derived from 17 distinct scenarios, mainly generated in Blender, where foreground masks are rendered together with RGB frames. The real-world subset is collected from diverse daily scenes, and we use SAM3~\cite{sam3} to extract foreground masks. These proxies cover a range of dynamic patterns, including rigid-body motion, soft-body deformation, object interaction, and fluid-like motion. We emphasize that this evaluation set is intended to diagnose whether a method can follow externally provided proxy dynamics while synthesizing prompt-aligned content, rather than to serve as a large-scale general physics benchmark.

\subsection{Evaluation Metrics}
We evaluate generated videos along four dimensions: visual quality, proxy-dynamics fidelity, interaction plausibility, and material/appearance consistency. Specifically, we adopt four relevant dimensions from VBench~\cite{huang2023vbench, huang2025vbench++, zheng2025vbench2}: Imaging Quality (IQ), Motion Rationality (MR), Mechanics (Mech.), and Material (Mat.). IQ measures visual distortions such as over-exposure, noise, and blur. MR evaluates whether the generated motion is temporally coherent and consistent with the motion pattern specified by the proxy video. Mech. focuses on the plausibility of object interactions, such as contact, collision, gravity-driven motion, and momentum transfer. Mat. measures temporal consistency and realistic rendering of materials. For MR, Mech., and Mat., similar to VBench, we annotate explicit descriptions and question-answering criteria conditioned on the proxy video and target prompt, and then perform video-based multi-question answering to assess whether the generated video follows the intended dynamics and interactions. We will release the detailed metrics and evaluation code after the review process.

\subsection{Baseline Implementations}
\label{sec:appendix_baseline_implementations}

For all baselines, we use their official implementations and run inference with the default hyperparameter configurations provided by the corresponding codebases. Specifically, Wan2.2 uses the T2V-A14B model~\cite{wan2025wan}; VACE uses its released 14B model trained for video inpainting~\cite{vace}; DiTFlow is implemented on CogVideoX-5B~\cite{yang2024cogvideox,ditflow}; and FlowDirector is implemented on Wan2.1-T2V-14B~\cite{li2025flowdirector,wan2025wan}. SDEdit is adapted to the same Wan2.2 backbone as ProxyUp. Numbers of their inference steps are 40 $\sim$ 50.

Because these methods are built on different base video generators and use different default inference schedules, fully aligning all experimental settings, especially the base model and the number of inference steps, is difficult. We therefore report the main comparison using each method's recommended default setting, and provide two additional controlled studies to better demonstrate the effect of the proposed SFR mechanism. First, we increase the number of inference steps used by SDEdit (Fig.~\ref{fig:sdedit_more_steps}), showing that the advantage of ProxyUp cannot be explained merely by using more sampling steps and instead comes from SFR. Second, we additionally evaluate cross-backbone variants, including ProxyUp on Wan2.1 (Fig.~\ref{fig:proxyup_wan21}) and DiTFlow on Wan2.2 (Fig.~\ref{fig:ditflow_wan22}), with result videos provided in the supplementary material. These studies help verify that ProxyUp's gains remain consistent when the backbone model is varied.

\section{Additional Qualitative Results}

\begin{figure}[tbp]
    \centering
    \includegraphics[width=\linewidth]{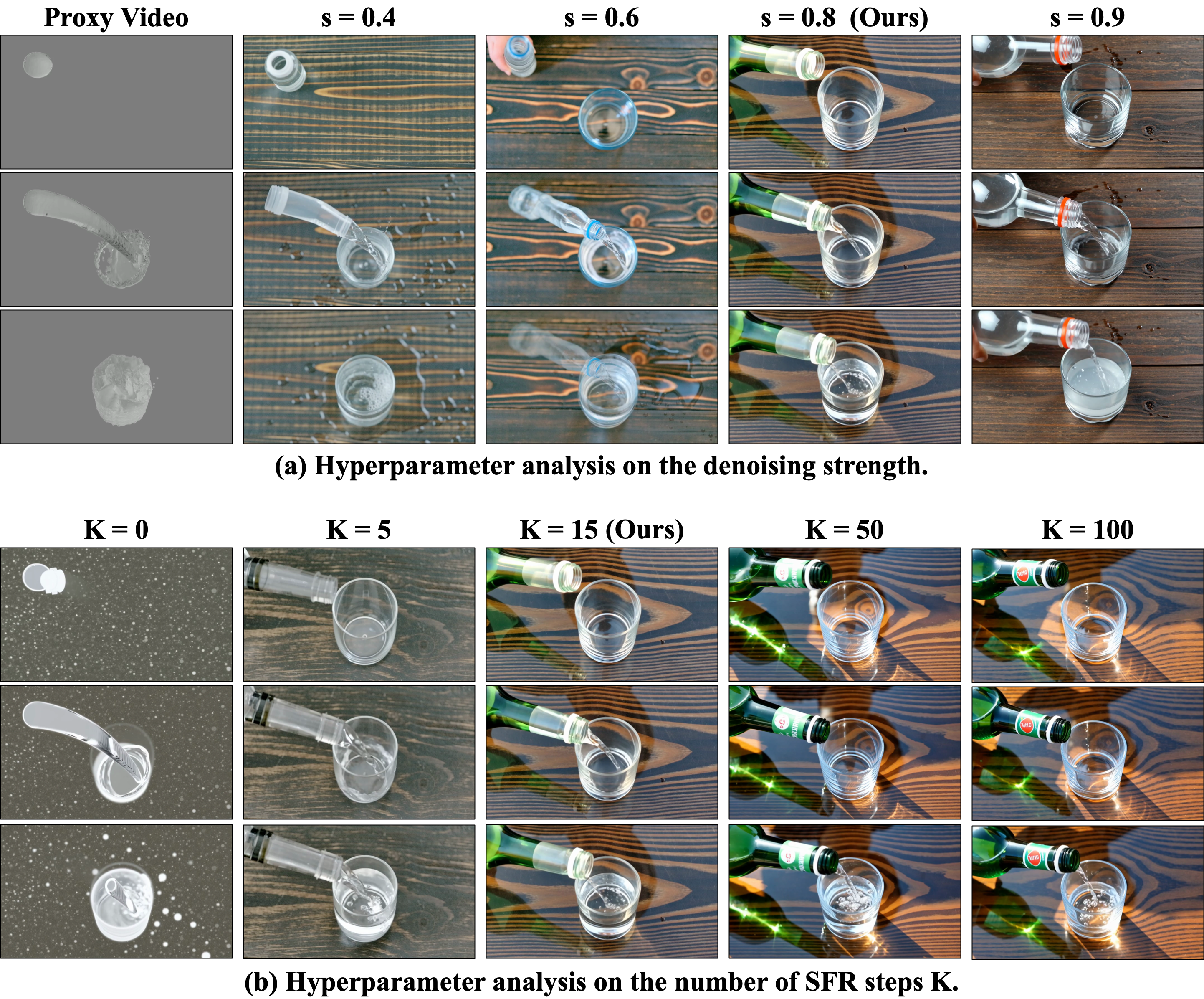}
    \caption{Hyperparameter analysis on the strength $s$ and the number of SFR steps $K$.}
    \label{fig:rs}
\end{figure}

\begin{figure}[tbp]
    \centering
    \includegraphics[width=\linewidth]{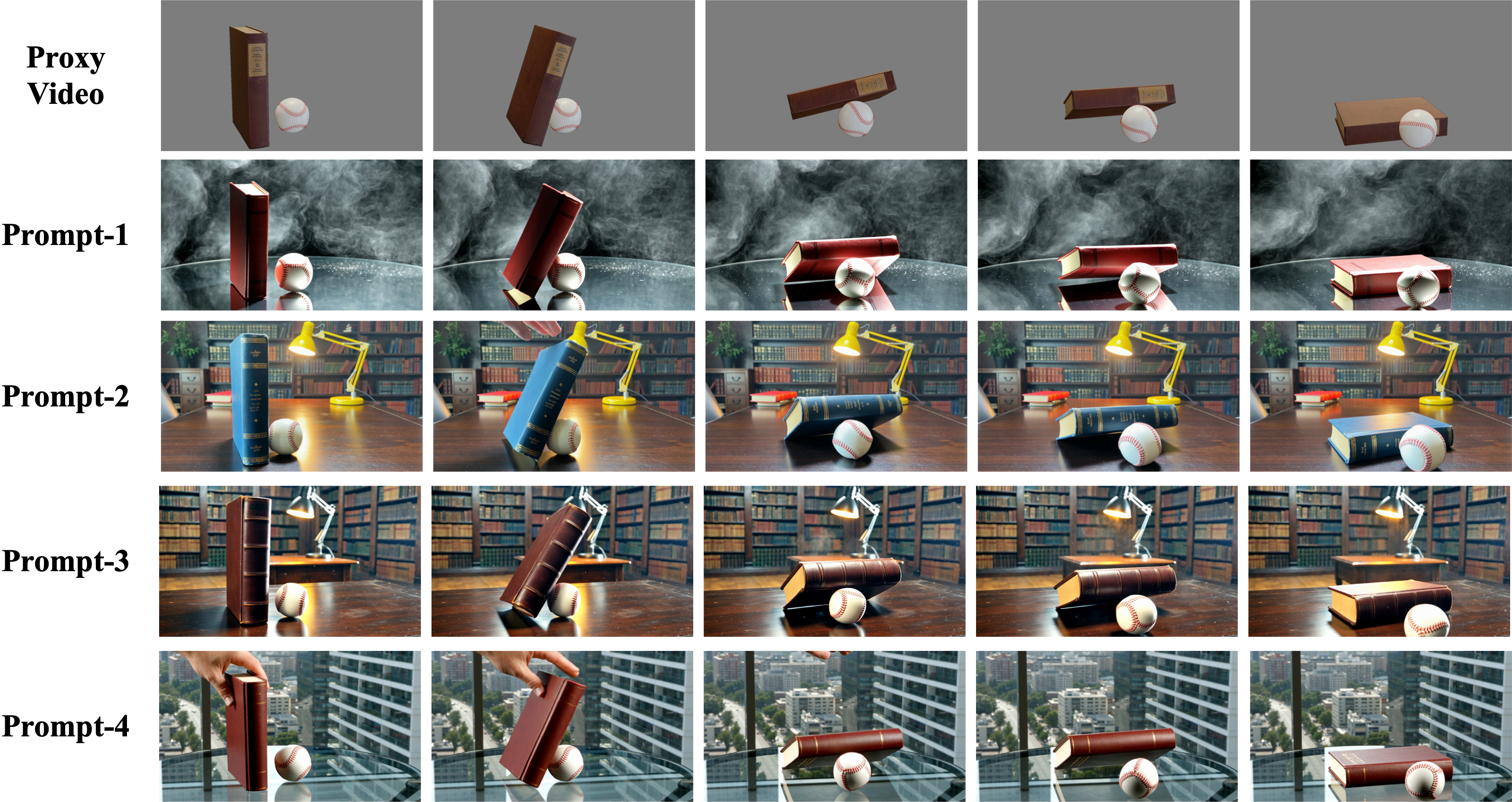}
    \caption{Given the same proxy video, ProxyUp can generate diverse videos conditioned on different prompts, highlighting its potential for data augmentation and simulation. Please refer to the supplementary video for the detailed prompts.}
    \label{fig:qua2}
\end{figure}

\subsection{Diverse Videos with Different Prompts}
In Fig.~\ref{fig:qua2}, we further show that, given a single proxy video, ProxyUp can generate diverse videos conditioned on different text prompts while preserving dynamics consistency. This highlights the potential of ProxyUp as a data augmentation tool for producing large amounts of physically consistent videos.

\subsection{SDEdit with More Inference Steps}

\begin{figure}[tbp]
    \centering
    \includegraphics[width=\linewidth]{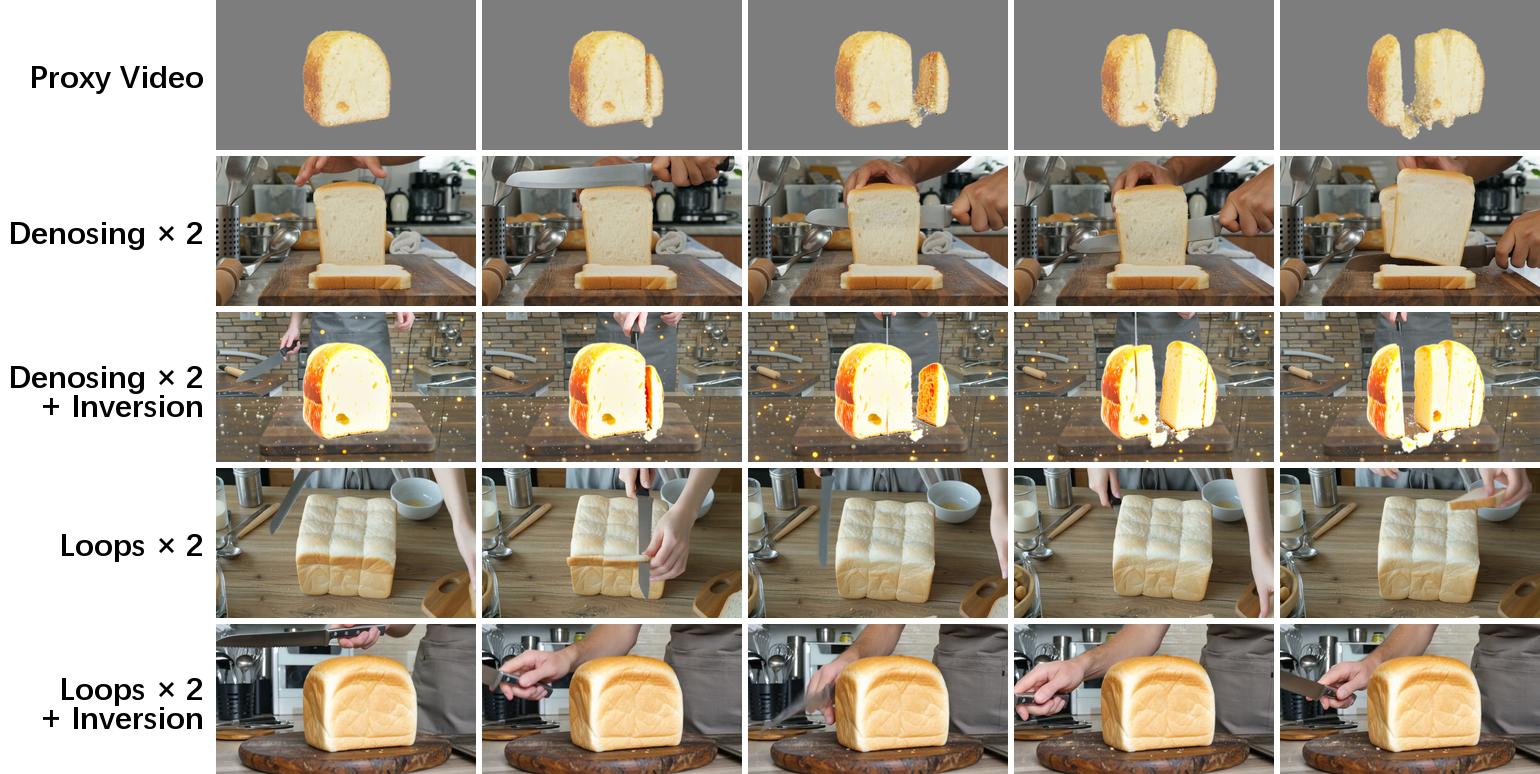}
    \caption{Qualitative results for increasing the inference budget of SDEdit. From top to bottom, the rows show the proxy video, SDEdit with doubled denoising steps, doubled denoising steps and inversion, two sampling loops, and two sampling loops with inversion.}
    \label{fig:sdedit_more_steps}
\end{figure}

Fig.~\ref{fig:sdedit_more_steps} evaluates whether the advantage of ProxyUp can be explained by simply allocating more sampling steps to SDEdit. Even with the increased inference budget (comparable or more than the number of inference steps in ProxyUp), SDEdit still struggles to preserve the proxy dynamics and synthesize coherent prompt-aligned content, supporting the importance of the proposed SFR mechanism. Here `Denoising $\times 2$' means the number of denoising steps in sampling is doubled. `Loops $\times 2$' means running the SDEdit 2 times, with the second loop using the output of the first loop as the initial latent. `Inversion' means applying inversion to the input latent before the SDEdit process, which is the same as the inversion step in ProxyUp. Note that our goal is to demonstrate the effect of the proposed SFR mechanism, thus we also apply region-wise latent noising (RLN) to SDEdit.

\subsection{ProxyUp on Wan2.1}

\begin{figure}[tbp]
    \centering
    \includegraphics[width=\linewidth]{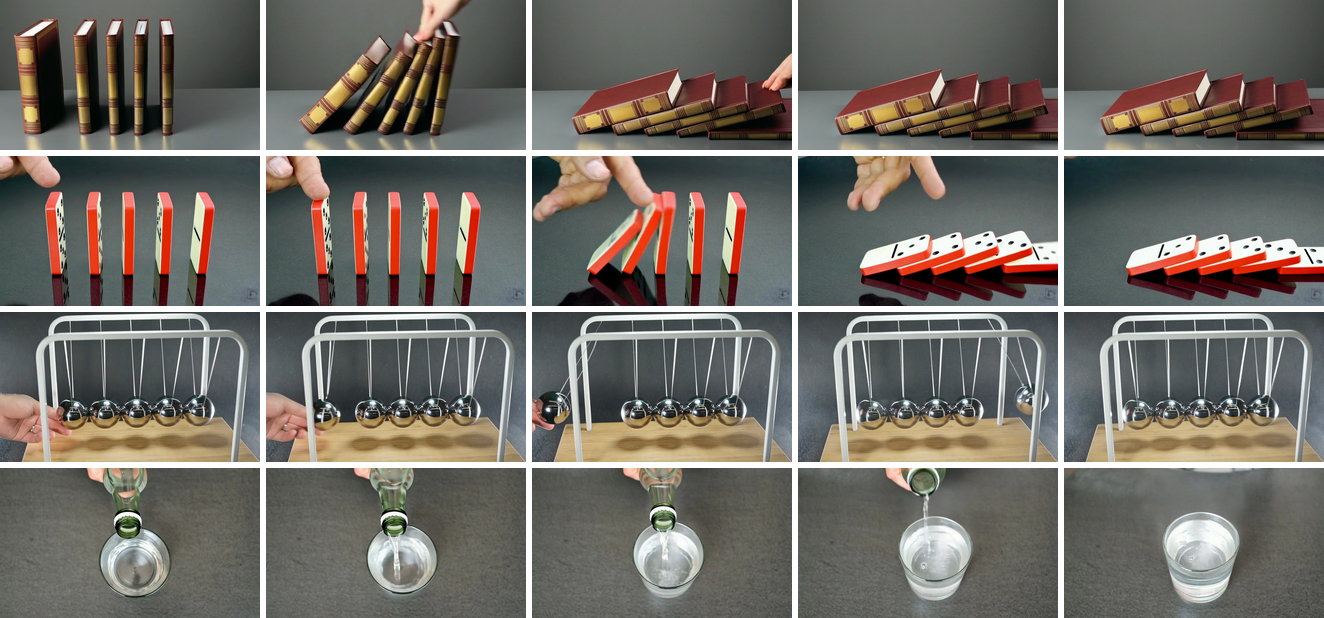}
    \caption{Cross-backbone qualitative results of ProxyUp on Wan2.1. The grid shows uniformly sampled frames from four representative generated videos.}
    \label{fig:proxyup_wan21}
\end{figure}

Fig.~\ref{fig:proxyup_wan21} shows that ProxyUp can be applied to Wan2.1 while still producing temporally consistent videos that follow the proxy dynamics. This cross-backbone result indicates that the benefits of ProxyUp are not tied to a single Wan backbone.

\subsection{DiTFlow on Wan2.2}

\begin{figure}[tbp]
    \centering
    \includegraphics[width=\linewidth]{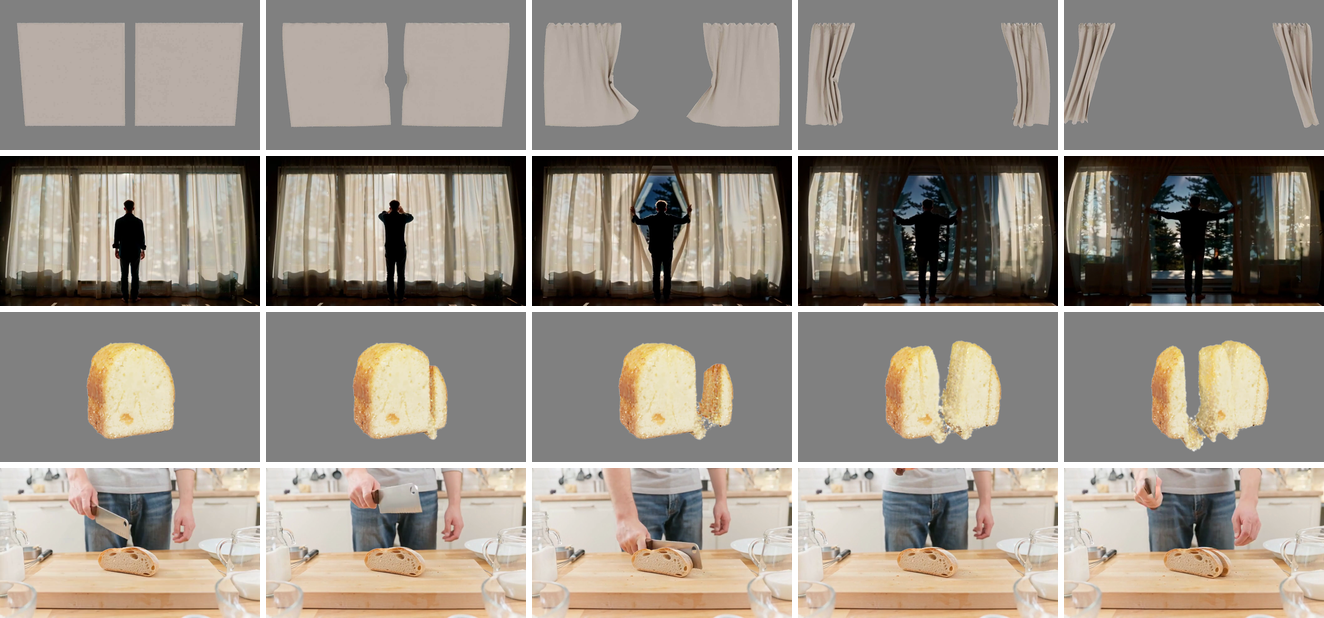}
    \caption{Cross-backbone qualitative results of DiTFlow on Wan2.2. Rows alternate between proxy videos and the corresponding DiTFlow generations on the Wan2.2 backbone.}
    \label{fig:ditflow_wan22}
\end{figure}

Fig.~\ref{fig:ditflow_wan22} provides additional cross-backbone examples for DiTFlow on Wan2.2. These results complement the ProxyUp-on-Wan2.1 study and help compare the behavior of different proxy-guided mechanisms under varied backbone settings.

\section{Failure Cases}
\label{sec:appendix_failure_cases}

\begin{figure}[tbp]
    \centering
    \includegraphics[width=\linewidth]{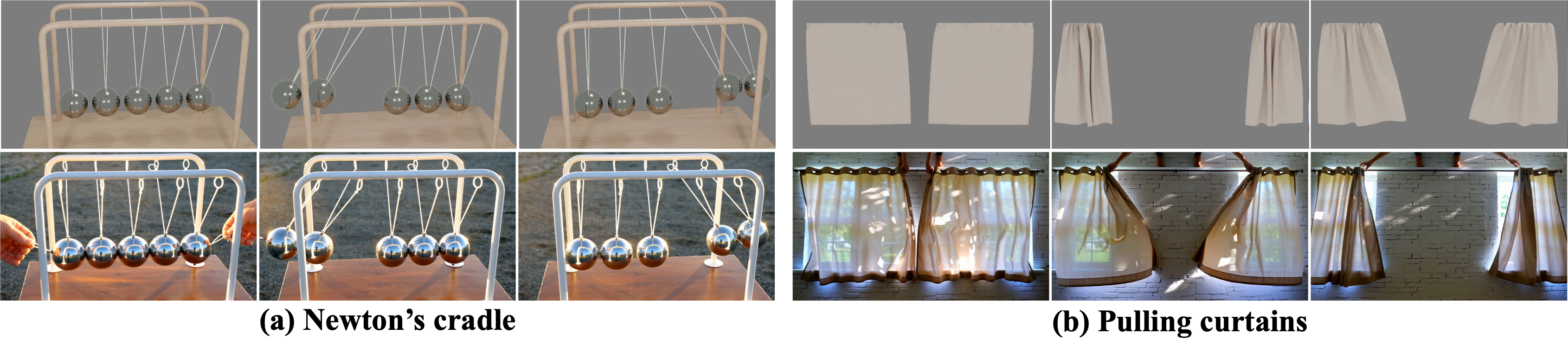}
    \caption{Failure cases of ProxyUp. When the proxy video contains uncommon or out-of-distribution dynamics, the video generation model may fail to correctly interpret the underlying interaction, leading to abnormal generation results.}
    \label{fig:fail}
\end{figure}

We present two representative failure cases in Fig.~\ref{fig:fail}. In the left example (Newton's cradle), the interaction between the person and the cradle is incorrect: the model fails to generate the action of lifting two balls. We attribute this failure to the rarity of such an action in the training data of current video generation models, which makes it difficult for the model to correctly understand and reproduce this behavior.

In the right example (pulling curtains), the person is generated above the curtain rod. This happens because the force application point in the proxy video is incorrectly placed at the top of the curtain. Interestingly, the generative model appears to recognize this issue and therefore places the person in a region that, while unrealistic, is still more consistent with the implied physical constraints. We hope these observations can further inspire future research on proxy-video-guided generation.



\end{document}